%% file: main.tex
\documentclass[runningheads]{llncs}

\usepackage{eccv}

\usepackage{eccvabbrv}   
\usepackage{graphicx}
\usepackage{booktabs}
\usepackage[accsupp]{axessibility}  
\usepackage{amsmath}
\usepackage{placeins}
\usepackage{float}
\usepackage{amssymb}
\usepackage{pifont}
\usepackage{float}      
\usepackage{placeins}   
\usepackage{tablefootnote}

\newcommand{\cmark}{\ding{51}}
\newcommand{\xmark}{\ding{55}}
\usepackage[pagebackref,breaklinks,colorlinks,citecolor=eccvblue]{hyperref}

\begin{document}
\raggedbottom

\title{Distilled Roads: Generalisable Road Network Extraction Across Sensors, Resolutions, and Regions}

\titlerunning{Distilled Roads}

\author{Sanayya \and Rakshith Sathish \and Ashwathi Nambiar}
\authorrunning{Sanayya et al.}
\institute{SatSure Analytics India Pvt. Ltd., Bengaluru, India}

\maketitle

\input{sec/0_abstract}
\input{sec/1_intro}

\input{sec/2_related}
\input{sec/3_problem}
\input{sec/4_method}
\input{sec/5_experiment}
\input{sec/6_results}

\input{sec/7_ablation}
\input{sec/8_failure}
\input{sec/9_conclusion}

\bibliographystyle{splncs04}
\bibliography{main}

\end{document}

%% file: sec/0_abstract.tex
\begin{abstract}
Road network segmentation from satellite imagery remains challenging due to large geographic variation in road appearance, occlusions, and domain shifts introduced by differing resolutions and sensors. Existing models, typically trained under narrow resolution--region combinations, generalise poorly to unseen environments such as rural settings, regions with distinct road materials, or imagery from new satellite platforms, often producing broken or disconnected predictions. Adapting these models to new domains usually requires retraining or fine-tuning, which is costly and risks catastrophic forgetting.

In this work, we reframe global road extraction as a continual adaptation problem rather than an architectural one. Our framework combines cross-resolution knowledge distillation across a resolution-decreasing curriculum, multi-sensor training, and topology-aware supervision, yielding a single model that generalises across $0.3-1.0$ m imagery from multiple satellite platforms across continents. On publicly available benchmarks, including City-Scale and Global-Scale, our model outperforms state-of-the-art results by up to $22$ F1 points and $15$ APLS points, while remaining the most efficient, with $3\times$ faster inference. Our results suggest that improved robustness across diverse sub-meter satellite imagery can be achieved through targeted training strategies, such as data curricula, distillation, and topology-aware losses, rather than increasingly complex architectures.

\keywords{Road Segmentation \and Knowledge Distillation \and Domain Generalisation \and Continual Learning \and Topology-Aware Supervision}
\end{abstract}

%% file: sec/1_intro.tex
\section{Introduction}
\label{sec:intro}

Accurate road network maps are essential for transportation planning, disaster response, and rural connectivity assessment, yet existing infrastructure data are often incomplete or outdated. Updating these maps typically requires manual cartographic work or ground surveys, which is time-consuming and costly at scale. To address this, several machine-learning models have been proposed \cite{samroad, 10716562} for identifying roads from satellite or aerial imagery. However, achieving reliable segmentation remains challenging as road appearances vary considerably across geographies, differ in width and surface material, and are frequently obscured by buildings, vegetation, or shadows. Further, these models are often trained for specific satellite sources, resolution or geographies. This limits the practical usage of such models for mapping at scale.
\begin{figure}[tb]
\centering
\begin{subfigure}[b]{0.48\linewidth}
    \includegraphics[width=\linewidth]{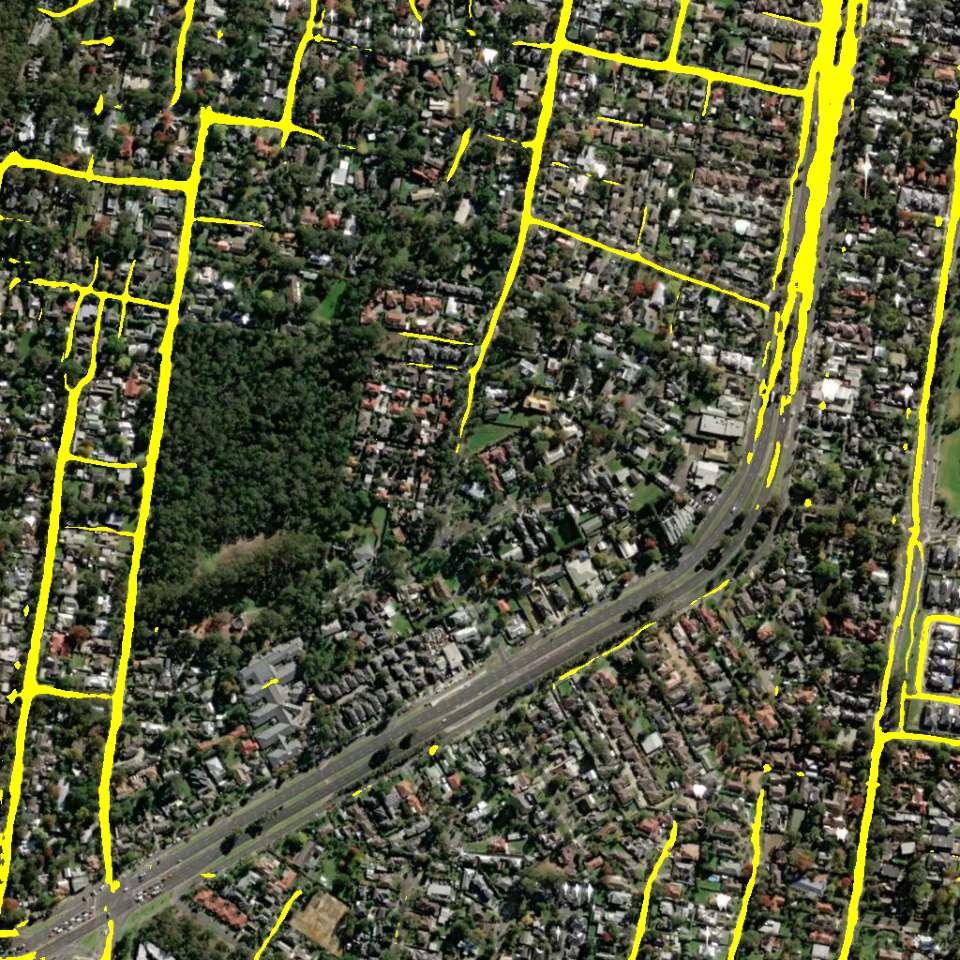}
    \caption{Baseline}
    \label{fig:conventional}
\end{subfigure}
\hfill
\begin{subfigure}[b]{0.48\linewidth}
    \includegraphics[width=\linewidth]{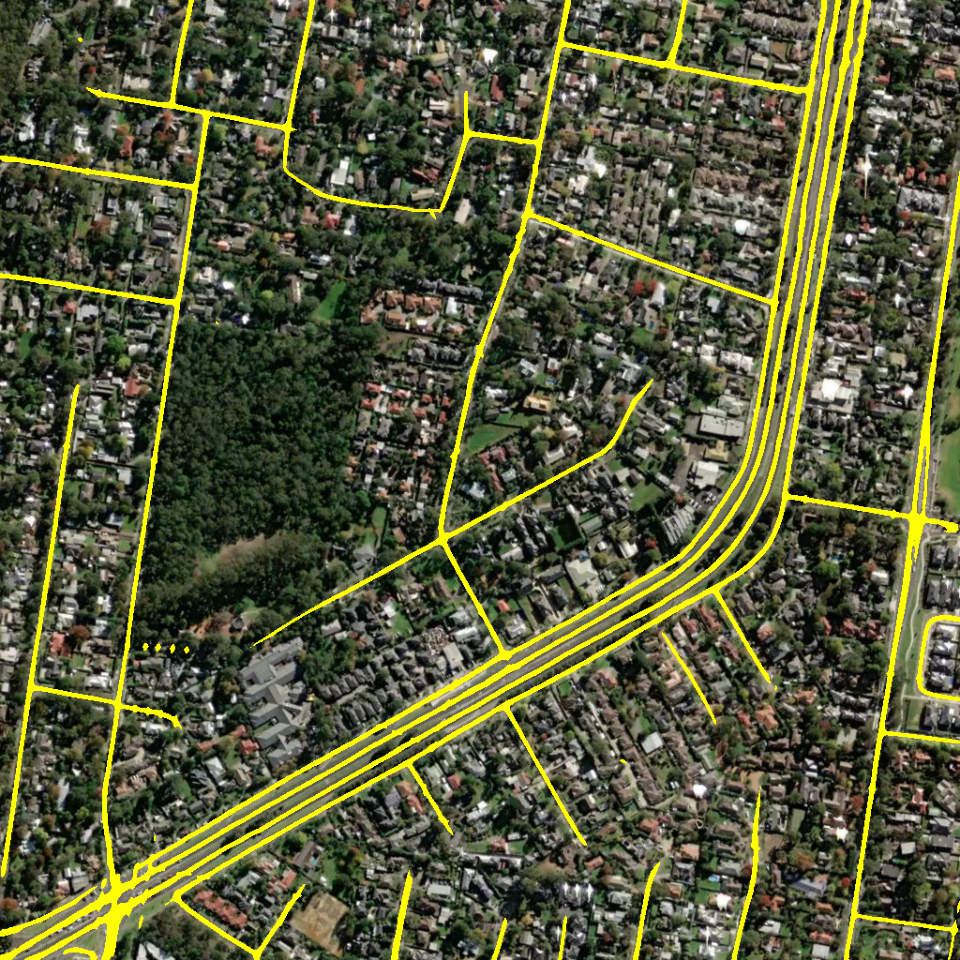}
    \caption{Ours}
    \label{fig:proposed}
\end{subfigure}
\caption{A comparison between a baseline model trained directly on multi-resolution, geographically diverse data without distillation or curriculum-based adaptation (a) and the proposed knowledge-distillation-based framework (b). The conventional approach exhibits fragmented predictions, whereas the proposed method produces more continuous and topologically coherent road networks across the scene.}
\label{fig:abstract}
\end{figure}


Prior works have shown that deep neural networks can identify roads from satellite imagery, but their performance deteriorates under changes in resolution, sensor characteristics, or geographic context. These failure modes suggest that the central limitation is robustness to domain shift, rather than a lack of architectural capacity, as incremental architectural modifications yield limited gains when the underlying data distribution changes significantly. Motivated by this, we focus on improving the training paradigm rather than introducing a new model design. Our approach targets the specific weaknesses of existing methods: poor generalisation across jointly varying resolutions, sensor characteristics, and geographic regions, and provides a framework that can be applied to standard segmentation architectures to improve their practical usability across diverse conditions.
To address these limitations, we introduce a curriculum-based knowledge-distillation framework in which a teacher trained on high-resolution single-domain imagery anchors fine-grained structural priors, and successive students inherit and preserve these priors via distillation while adapting to progressively broader, lower-resolution domains. This formulation encourages the student to learn representations that remain consistent across varying spatial resolutions and sensor characteristics, enabling the model to generalise more reliably to new geographies. The framework integrates three complementary supervisory signals: (i) a distillation loss that aligns the student's outputs with the teacher's soft predictions; (ii) a topology-aware loss that promotes continuity in thin road structures; and (iii) standard semantic segmentation losses to stabilise optimisation.

%% file: sec/2_related.tex
\section{Related Works}
\label{sec:related}

Road network extraction from overhead imagery has been a subject of study for decades. Early systems relied on hand-crafted features or graphical reasoning, such as CRF-based connectivity modelling \cite{WEGNER2015128,
HINZ200383}. With the rise of deep learning, CNN-based segmentation has become the dominant approach in image segmentation. Patch or pixel classifiers followed by thinning \cite{Mnih} improve local segmentation yet often struggle to preserve long-range topology.
Graph-based methods aim to directly extract road structure. RoadTracer \cite{road_tracer}incrementally expands a graph, whereas Sat2Graph\cite{sat2graph}introduces a hybrid graph-tensor representation for end-to-end learning. These approaches improve structure but remain sensitive to distribution shifts and scale inconsistencies. The DeepGlobe 2018 challenge \cite{deepglobe} established D-LinkNet \cite{dlink} as a widely adopted encoder--decoder network for road extraction. Follow-up models incorporate stronger multi-scale reasoning using dilated convolutions, or deformable features \cite{chen2018encoder, wang}. However, most optimise only pixel metrics, leaving topology restoration to post-processing. Transformer-based methods, such as RNGDet++ \cite{xu2023rngdet++} and SAM-Road \cite{samroad}, demonstrate strong large-scale performance; however, they rely on complex, multi-stage pipelines that limit their deployability.

\noindent\textbf{Generalisation across regions and sensors:} Road appearance varies with geography, season, and acquisition platform. UDA frameworks, such as RoadDA \cite{roadda}, align distributions via GAN-based adaptation and pseudo-label refinement. Foundation-style models, such as CrossEarth\cite{gong2024crossearth}, enhance cross-domain transfer but require very large backbones and substantial training compute.

\noindent\textbf{Knowledge distillation for continual updates:} Knowledge distillation (KD) is a well-established technique for transferring knowledge from a teacher model to a student, typically by having the student mimic the teacher's soft output distributions \cite{hinton2015distilling}. KD is now widely used in continual semantic segmentation \cite{yangze} to mitigate catastrophic forgetting by preserving soft teacher predictions. Recently, KD has also been shown to be effective in continual-learning settings beyond class segmentation, \eg, neural-field domains with varying signal distributions \cite{visser2025knowledge}.

%% file: sec/3_problem.tex
\section{Problem Statement}
\label{sec:problem}

We address the task of road-network segmentation from satellite imagery, formulated as a binary semantic segmentation problem. Given an input image
\( I \in \mathbb{R}^{H \times W \times 3} \), the objective is to predict a mask
\( \hat{Y} \in \{0,1\}^{H \times W} \) identifying road and background pixels, while preserving the continuity of the underlying road network.

Each image is associated with a spatial resolution
\( r \in [0.3, 1] \) m, a sensor identity
\( s \in \mathcal{S} \) (\eg, WorldView-3, GeoEye), and a geographic region
\( g \in \mathcal{G} \) spanning diverse urban and rural environments.
We aim to learn a single segmentation function
\[
f_{\theta}: I \rightarrow \hat{Y},
\]
that generalises across these domains without retraining when the resolution, sensor, or location changes.

%% file: sec/4_method.tex
\section{Method}

\begin{figure}[tb]
    \centering
    \includegraphics[width=\linewidth]{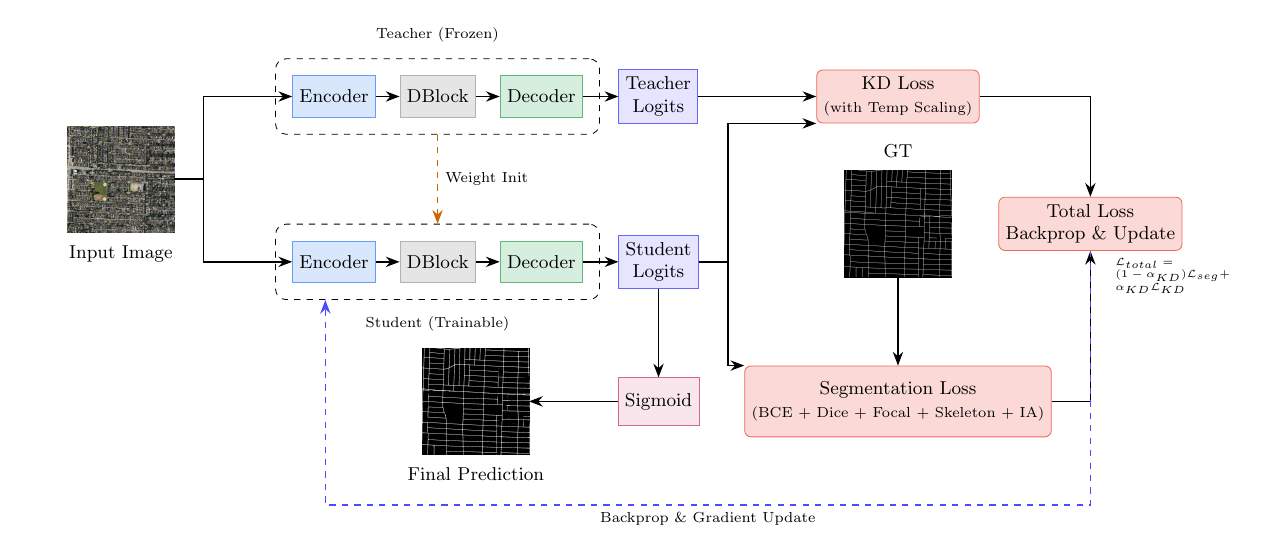}
    \caption{\textbf{Overview of the proposed framework}. We propose a teacher--student knowledge-distillation paradigm in which a frozen previous-stage teacher, initially trained on high-resolution imagery, provides soft supervisory signals to guide a trainable student on lower-resolution, broader-domain data. Both networks share the same encoder--decoder architecture with Dilation Block modules. Given an input image, the teacher produces temperature-scaled logits and compares them with the student's logits through a distillation loss. In parallel, the student receives supervision from ground-truth labels via a composite segmentation loss consisting of Binary Cross-Entropy, Dice, Focal, Skeleton, and Intersection-Aware terms. The total loss combines these objectives according to a weighting parameter, allowing the student to inherit structural priors from the teacher while learning to adapt to new domains. After backpropagation, only the student model is updated; the teacher remains fixed. The final prediction is obtained by applying a sigmoid activation to the student logits.}
    \label{fig:method}
\end{figure}
The overall architecture of the proposed framework is illustrated in \cref{fig:method}.

\subsection{Segmentation Network}
Our core segmentation backbone follows the general encoder--decoder design, inspired by DLinkNet-34 \cite{dlink}, which has proven effective in capturing thin, elongated, and topologically structured patterns. It consists of three main components: a hierarchical encoder, a dilation block, and a decoder equipped with skip connections. A pretrained ResNet-34 is used as the backbone architecture of the encoder. Roads exhibit substantial variability across geography, ranging from narrow agricultural tracks to wide multilane highways. A single receptive field cannot capture this diversity, particularly when occlusions, changing illumination or when surrounding objects appear similar.
To address this, the encoder in DLinkNet-34 is paired with a dilation block comprising multiple dilated convolutions with dilation rates $D\in\{1,2,4,8\}$, which progressively increase the receptive field (\eg, $3\times3$, $7\times7$, $15\times15$, and $31\times31$) while preserving the spatial resolution.

\subsection{Curriculum-Based Cross-Resolution Distillation}
\label{sec:kd}

Adapting a road-segmentation model to new geographies, resolutions, and sensors
tends to erode the structural priors learned from earlier data, a form of
catastrophic forgetting. We address this by coupling knowledge distillation (KD)
with a staged training curriculum that anchors high-resolution structural priors
from a controlled single-domain teacher and then progressively broadens
geographic and sensor diversity while decreasing spatial resolution.

\paragraph{\textbf{Staged curriculum with a rolling teacher.}}
The curriculum proceeds through a sequence of stages of decreasing spatial
resolution and increasing domain diversity. The first stage trains a plain semantic-segmentation network on the aerial corpus; this model anchors fine-grained structural priors derived from high-resolution imagery. At every subsequent stage, the model trained on the preceding stage is frozen and serves as the \emph{teacher}, while a \emph{student}, initialised from the teacher's weights, is trained on the new stage's data under joint supervision from its ground-truth labels and the teacher's softened predictions. Because resolution decreases monotonically across the curriculum, each teacher was trained at a resolution equal to or higher than the resolution the student now learns. Distillation, therefore, transfers structurally richer, higher-resolution priors, while weight initialisation and the distillation term jointly preserve the priors accumulated over earlier stages. We refer to this, a teacher carrying higher-resolution-derived priors supervising a student on lower-resolution data, as \emph{cross-resolution distillation}; it is a property
of the curriculum rather than of paired high/low-resolution inputs, as teacher and student process the same (current-stage) imagery at distillation time. The framework is defined for an arbitrary number of sequential stages: each newly trained student can, in turn, serve as the frozen teacher for the next stage, thereby progressively accumulating structural priors. In this work, we instantiate a single distillation step—a teacher trained on the high-resolution aerial corpus supervising a student trained on the lower-resolution Global-Scale data to demonstrate cross-resolution transfer; extension to deeper curricula is left for future work.
\paragraph{\textbf{Distillation objective.}}
Let $z_t$ and $z_s$ denote the teacher and student logits for a given input. We
soften both with a temperature $T>1$ and align them through a mean-squared-error
distillation loss:
\begin{equation}
\mathcal{L}_{KD} = \mathrm{MSE}\!\left(\sigma\!\left(\tfrac{z_s}{T}\right),\,
\sigma\!\left(\tfrac{z_t}{T}\right)\right) T^2,
\end{equation}
where $\sigma(\cdot)$ is the sigmoid activation. We use MSE instead of KL divergence because sigmoid-based binary predictions yielded more stable optimisation in our setting. The softened targets provide smoother supervisory signals that capture uncertainty near road boundaries and
reduce sensitivity to local appearance variation, constraining the student to
remain close to the teacher's output distribution while adapting to the new
domain. The $T^2$ factor follows standard temperature-scaled distillation
practice~\cite{hinton2015distilling}, retained to keep the distillation term on a scale
comparable to the segmentation loss. The student is optimised with the composite
segmentation loss $\mathcal{L}_{\mathrm{Seg}}$ of Sec.~\ref{sec:loss}
(Eq.~\ref{eq:final_loss}, evaluated against ground truth) combined with the distillation
term through a single mixing weight $\alpha_{KD}\in[0,1]$:
\begin{equation}
\mathcal{L}_{\mathrm{total}} = (1-\alpha_{KD})\,\mathcal{L}_{\mathrm{Seg}}
+ \alpha_{KD}\,\mathcal{L}_{KD}.
\end{equation}
Only the student is updated at each stage; the teacher remains fixed. 

\paragraph{\textbf{Role of distillation.}}
The objective is designed to act in three ways. By anchoring the student to the frozen teacher, it is intended to preserve fine-grained structural cues established at earlier, higher-resolution stages that adaptation would otherwise overwrite; to limit large shifts in the encoder representation during broader-domain fine-tuning, mitigating catastrophic forgetting, and to let the student integrate new-domain information while remaining aligned with the teacher's distribution, yielding more stable adaptation across regions and sensors. We evaluate the extent to which each effect holds in Sec .~\ref {sec:ablation}.

\subsection{Loss Function Design}\label{sec:loss}

To supervise the student model, we use a composite loss ($\mathcal{L}_{\mathrm{Seg}}$) that combines complementary objectives, each targeting a different aspect of the road-segmentation task. To provide stable pixel-wise gradients, we use binary cross entropy ($\mathcal{L}_{\mathrm{BCE}}$) as shown in \cref{eq:bce}.

\begin{equation}
\mathcal{L}_{\mathrm{BCE}} = - \left[ G \log(\sigma(z)) + (1 - G) \log(1 - \sigma(z)) \right].
\label{eq:bce}
\end{equation}

\noindent where \(z\), \(\sigma(z)\), and \(G\) represent raw model logits, sigmoid activation, and ground-truth mask, respectively.

To compensate for $\mathcal{L}_{\mathrm{BCE}}$'s sensitivity to foreground imbalance and its tendency to miss thin structures, we introduce dice loss ($\mathcal{L}_{\mathrm{Dice}}$) as defined in \cref{eq:dice}

\begin{equation}
\mathcal{L}_{\mathrm{Dice}} = 1 -
\frac{2 \sum (P \cdot G) + \epsilon}{\sum P + \sum G + \epsilon}.
\label{eq:dice}
\end{equation}

\noindent where \(P = \sigma(z)\), \(G\), and \(\epsilon\) represent predicted probability, ground-truth mask and smoothing constant, respectively.

Even with the inclusion of Dice loss, the model remained sensitive to occlusions and variations in illumination. To address this limitation, we incorporate focal loss ($\mathcal{L}_{\mathrm{Focal}}$), as formulated in \cref{eq:focal}, which downweights easy examples and amplifies the contribution of challenging ones.

\begin{equation}
\mathcal{L}_{\mathrm{Focal}} = \alpha (1 - p_t)^{\gamma} \mathcal{L}_{\mathrm{BCE}},
\qquad p_t = e^{-\mathcal{L}_{\mathrm{BCE}}}.
\label{eq:focal}
\end{equation}

\noindent where \(\alpha\) and \(\gamma\) represents weighting factor and focusing parameter.

In addition to the pixel-wise losses, we introduced a skeleton recall loss \cite{kirchhoff2024skeleton} to mitigate small $1$--$2$ px discontinuities observed along elongated road segments. The ground-truth mask is thinned via morphological skeletonisation to obtain a one-pixel-wide centerline structure. We define $\mathcal{L}_{\mathrm{Skel}}$ as in \cref{eq:skel}

\begin{equation}
\mathcal{L}_{\mathrm{Skel}} = 1 -
\frac{\sum (P \cdot S) + \epsilon}{\sum S + \epsilon}.
\label{eq:skel}
\end{equation}

\noindent where $P$ denotes the predicted probability map, and $S$ denotes the centerline structure. This penalises missing predictions along the skeleton, effectively measuring the recall of structural centerlines.
Further, to reduce the disconnectivity observed at road intersections (T-junctions, Y-splits), we extended $\mathcal{L}_{\mathrm{Skel}}$ with an intersection-aware enhancement. Pixels with 3 or more neighbouring pixels in the centerline structure (S) are considered to be intersection points. The intersection loss ($\mathcal{L}_{\mathrm{int}}$) is formulated as in \cref{eq:int}.

\begin{equation}
\mathcal{L}_{\mathrm{Int}} =
\frac{\sum J \cdot (1 - P)}{\sum J + \epsilon}.
\label{eq:int}
\end{equation}

\noindent where \(J\) represents intersection pixels (skeleton points with $\geq 3$ neighbors) and \(P\) represents predicted probability.

Hence, the final training objective ($\mathcal{L}_{\mathrm{Seg}}$) is as follows:

\begin{equation}\label{eq:final_loss}
  \mathcal{L}_{\mathrm{Seg}} = \mathcal{L}_{\mathrm{BCE}} + \mathcal{L}_{\mathrm{Dice}} + \mathcal{L}_{\mathrm{Focal}} + \mathcal{L}_{\mathrm{Skel}} + \mathcal{L}_{\mathrm{Int}}
\end{equation} 
\FloatBarrier

%% file: sec/5_experiment.tex
\section{Experiments}
\label{sec:experiments}

\noindent\textbf{Datasets:} Robust road extraction across continents, sensors, and resolutions requires datasets that span diverse geographic and imaging conditions. Road morphology varies globally due to differences in materials, climate, and surrounding land cover, and satellite sensors exhibit distinct radiometric and spectral characteristics. Most public datasets cover limited regions, typically a single country or a few cities, and are therefore inadequate for training models intended to generalise across diverse global regions.

While the distillation framework supports a multi-stage curriculum of decreasing resolution and increasing diversity, we instantiate it here with a single round of distillation: a high-resolution single-region aerial teacher ($0.3-0.5 m)$ and a student trained on globally distributed Global-Scale imagery ($1.0 m$). DeepGlobe-pretrained weights are used only to initialise the teacher. The first stage anchors fine-grained structural priors using a high-resolution, single-region aerial corpus; the second broadens geographic and sensor coverage with globally distributed imagery at coarser resolution. At each stage, the model trained on the preceding stage is frozen and serves as the teacher for the next, so that high-resolution-derived priors are progressively carried forward and preserved while the student adapts to broader, lower-resolution domains. Table~\ref{tab:datasets} summarises the datasets used in this work. 

Evaluation is conducted under three settings: (i) Global-Scale (In-Domain), using the Global-Scale dataset from Table~\ref{tab:datasets} with images drawn from the training distribution; (ii) Global-Scale (Out-of-Domain), using the same dataset but from unseen regions; and (iii) City-Scale, an independent road segmentation benchmark used only for evaluation.

\begin{table}[H]
\centering
\caption{Datasets used in this work. Training indicates whether the dataset was used during model training (\cmark) or reserved for evaluation only (\xmark).}
\label{tab:datasets}
\resizebox{\linewidth}{!}{
\begin{tabular}{lccccc}
\toprule
\textbf{Dataset} & \textbf{Training} & \textbf{Resolution} & \textbf{Sensor(s)} & \textbf{Location} & \textbf{Total Area (km$^2$)}\\
\midrule
DeepGlobe
& \xmark \tablefootnote{Model trained on DeepGlobe is used for weight initialisation, dataset is not used for training.}
& 0.5 m
& WorldView-3, WorldView-2, GeoEye-1
& Thailand, Indonesia, India
& 2,220 \\
Proprietary Aerial Dataset
& \cmark
& 0.3--0.5 m
& Various aerial sensors
& India
& 400 \\
Global-Scale (train set)
& \cmark
& 1.0 m
& Commercial satellites
& Global (except Antarctica)
& 10,000 \\
\midrule
Global-Scale (out of domain)
& \xmark
& 1.0 m
& Commercial satellites
& Global
& 545 \\
Global-Scale (in domain)
& \xmark
& 1.0 m
& Commercial satellites
& Global
& 2,600 \\
City-Scale
& \xmark
& 1.0 m
& Google Static Maps API
& 20 U.S.\ cities
& 720 \\
\bottomrule
\end{tabular}}
\end{table}
\FloatBarrier

\noindent\textbf{Data Preprocessing:} To ensure that imagery from different sensors, resolutions, and acquisition platforms can be used within a single unified model, we apply a consistent preprocessing pipeline across all datasets. All images are  converted to a consistent colour representation and normalised using an affine transform that recenters pixel distributions into a stable range:

\begin{equation}
I_{\mathrm{norm}} = \frac{I}{255} \times 3.2 - 1.6.
\end{equation}
\noindent where \(I\) denotes the input image.

All data sources are processed at their native resolution and subsequently tiled into $1024 \times 1024$ patches. Furthermore, all road annotations are converted into a unified binary road footprint mask. This standardisation resolves inconsistencies between datasets with differing annotation styles, such as thin centerlines versus polygonal footprints.

\noindent\textbf{Data Augmentations:} To enhance robustness to appearance shifts across continents, we employ a domain-randomisation augmentation pipeline drawing on strategies used in SAM++ \cite{sam++} and DINO \cite{Caron_2021_ICCV}. The goal is to expose the model to distributional variations observed across continents, sensors, and environmental conditions.
The augmentations applied during training include:
\begin{itemize}
    \item \textit{Geometric}: Random rotations, flips, elastic deformations, and scale-rotation transformations
    \item \textit{Photometric}: Brightness and contrast adjustments, hue--saturation jitter, and synthetic shadow insertion
    \item \textit{Atmospheric}: Fog and haze synthesis to simulate adverse weather conditions
    \item \textit{Sensor-domain perturbations}: Blur, additive noise, and compression to emulate differences across aerial and satellite platforms
\end{itemize}

\noindent\textbf{Data Split:} To ensure that evaluation reflects true generalisation rather than memorisation of spatially adjacent regions, we adopt a leakage-free data-splitting strategy. We therefore partition each dataset at the regional level, assigning entire districts, urban areas, or rural zones exclusively to either the training or validation set so that no contiguous areas appear in multiple splits. For multi-sensor datasets, imagery from different acquisition platforms remains naturally separated, avoiding cross-domain contamination. For the Global-Scale dataset, we further separate the training and evaluation regions at the country or continental scale to assess cross-region generalisation, a central objective of this work.

\noindent\textbf{Evaluation Metrics:} We evaluate model performance using a combination of pixel-level and topology-aware metrics that together capture both the local accuracy of road extraction and the global connectivity of the reconstructed network. As most datasets annotate roads as thin centreline structures, whereas the model predicts full-width road surfaces, strict pixel-level matching would disproportionately penalise minor thickness discrepancies. To account for this, we apply an $8$-connected morphological dilation to the ground-truth centreline with a tolerance radius of $r=3$ pixels, and consider a predicted pixel correct if it falls within the dilated region. This tolerance-based matching provides a more appropriate evaluation for full-width predictions derived from centreline annotations.

Formally, let $\mathrm{Dilate}(G, r)$ denote the morphological dilation of the ground-truth mask $G$ by radius $r$ and let $P$ denote the binarized road prediction. True positives (TP), false positives (FP), and false negatives (FN) are defined as:
\begin{equation}
\mathrm{TP} = | P \cap \mathrm{Dilate}(G, r) |
\end{equation}
\begin{equation}
\mathrm{FP} = | P \setminus \mathrm{Dilate}(G, r) |
\end{equation}
\begin{equation}
    \mathrm{FN} = | G \setminus \mathrm{Dilate}(P, r) |
\end{equation}

\noindent These are used to calculate tolerance-based precision, recall, and F1-score. Tolerance-based precision, recall, and F1 are computed from these definitions, with F1 serving as the primary segmentation metric.

To evaluate the connectivity and usability of the predicted road network, we also compute the Average Path Length Similarity (APLS) \cite{van2019city}. This metric compares the predicted and ground-truth graphs by measuring how similarly they connect junctions and support routing. After skeletonising both masks into graph structures, we sample node pairs $(u,v)$ and compute their shortest-path distances in the ground-truth graph $d_{\mathrm{GT}}(u,v)$ and the
predicted graph $d_{\mathrm{PR}}(u,v)$. The metric is defined as:

\begin{equation}
\mathrm{APLS} = 1 - \frac{1}{N}
\sum_{(u,v)}
\frac{\left| d_{\mathrm{GT}}(u,v) - d_{\mathrm{PR}}(u,v) \right|}
{d_{\mathrm{GT}}(u,v) + \epsilon},
\end{equation}
\noindent where $N$ is the number of sampled node pairs and $\epsilon$ is a small constant to avoid division by zero. APLS approaches a value of $1$ when the predicted network preserves junction topology, connectivity, and realistic path lengths, and decreases when roads are fragmented or misaligned.

\noindent\textbf{Baselines:} We compare our method against two recent state-of-the-art road extraction approaches, SAM-Road \cite{samroad} and RNGDet++ \cite{xu2023rngdet++}, which represent graph-based and transformer-based pipelines designed specifically for large-scale road network extraction. These methods incorporate complex multi-stage processing, providing strong and competitive baselines for evaluating cross-domain generalisation. All methods are evaluated using the same metric implementation and test imagery to ensure a fair comparison.

%% file: sec/6_results.tex
\section{Results and Discussion}
\label{sec:results}

A comparison of the proposed method with baselines is provided in \cref{tab:benchmark}. It is worth noting that the proposed model achieves the optimal balance of pixel accuracy and topological correctness across global datasets. The model achieves the highest F1 score ($86.02$) and strong connectivity (APLS = $68.55$) on the `in-domain' set of the Global-Scale dataset, representing a substantial improvement over prior segmentation-based approaches. More importantly, the model generalises well on the `out-of-domain' set of the Global-Scale dataset that was not included during training: we again achieve the highest F1 ($74.47$) and APLS ($55.22$). In the City-Scale dataset, the model achieves the highest pixel-level accuracy (F1 = $85.69$) while maintaining the highest connectivity (APLS = $ 83.16$). Notably, the City-Scale results are obtained under different training conditions for each method. SAM-Road and RNGDet++ are trained on the City-Scale training split and evaluated in-domain, whereas our model is never exposed to City-Scale during training; it is applied zero-shot, having been trained only on DeepGlobe, the aerial corpus, and the Global-Scale dataset.

\begin{table}[H]
\centering
\caption{Comparison with state-of-the-art road extraction methods on three evaluation settings.}
\label{tab:benchmark}
\resizebox{\linewidth}{!}{
\begin{tabular}{lcccccccccccccccc}
\toprule
 & \multicolumn{4}{c}{\textbf{Global-Scale (In-Domain)}}
 & \multicolumn{4}{c}{\textbf{Global-Scale (Out-of-Domain)}}
 & \multicolumn{4}{c}{\textbf{City-Scale}}\\
\textbf{Name} & \textbf{Pr} & \textbf{Re} & \textbf{F1} & \textbf{APLS}
              & \textbf{Pr} & \textbf{Re} & \textbf{F1} & \textbf{APLS}
              & \textbf{Pr} & \textbf{Re} & \textbf{F1} & \textbf{APLS}\\
\midrule
RNGDet++
& 65.31 & 62.25 & 63.74 & 52.72
& 54.22 & 53.08 & 53.65 & 38.08
& 80.07 & 89.36 & 84.46 & 63.14 \\

SAM-Road
& 83.87 & 45.33 & 58.85 & 59.08
& \textbf{75.76} & 47.20 & 58.16 & 40.51
& 80.33 & 84.75 & 82.48 & 68.37\\

\textbf{Ours}
& \textbf{84.68} & \textbf{87.40} & \textbf{86.02} & \textbf{68.55}
& 73.19 & \textbf{75.79} & \textbf{74.47} & \textbf{55.22}
& \textbf{82.20} & \textbf{89.48} & \textbf{85.69} & \textbf{83.16}\\
\bottomrule
\end{tabular}}
\end{table}

\begin{figure}[H]
\centering

\begin{minipage}{0.23\linewidth}
\centering
\textbf{Ground Truth}\\[2pt]
\includegraphics[width=\linewidth]{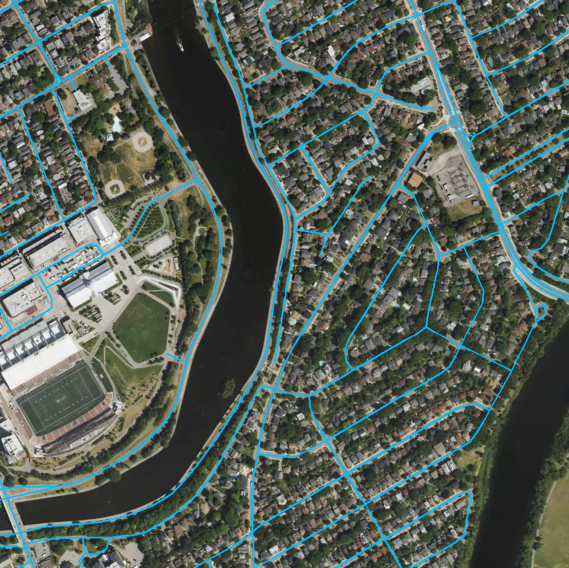}
\end{minipage}
\begin{minipage}{0.23\linewidth}
\centering
\textbf{RNGDet++}\\[2pt]
\includegraphics[width=\linewidth]{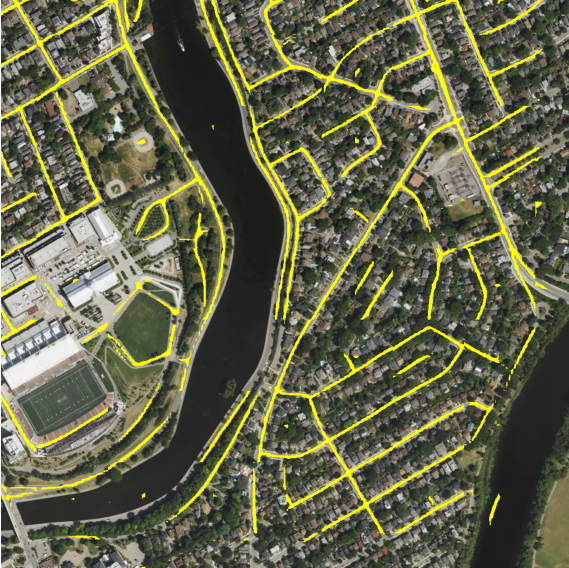}
\end{minipage}
\begin{minipage}{0.23\linewidth}
\centering
\textbf{SAM-Road}\\[2pt]
\includegraphics[width=\linewidth]{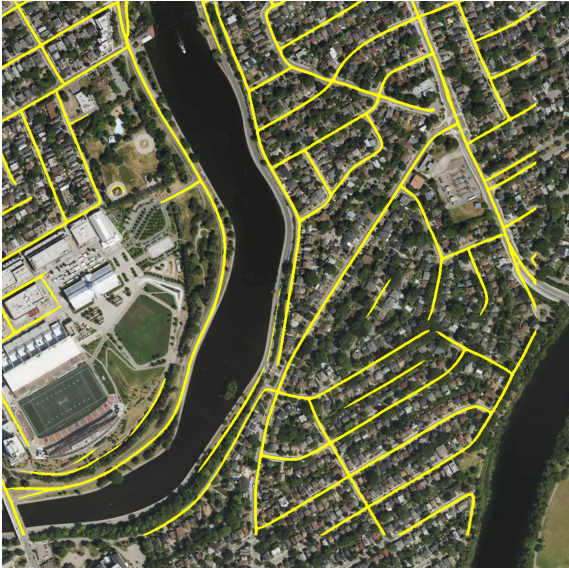}
\end{minipage}
\begin{minipage}{0.23\linewidth}
\centering
\textbf{Ours}\\[2pt]
\includegraphics[width=\linewidth]{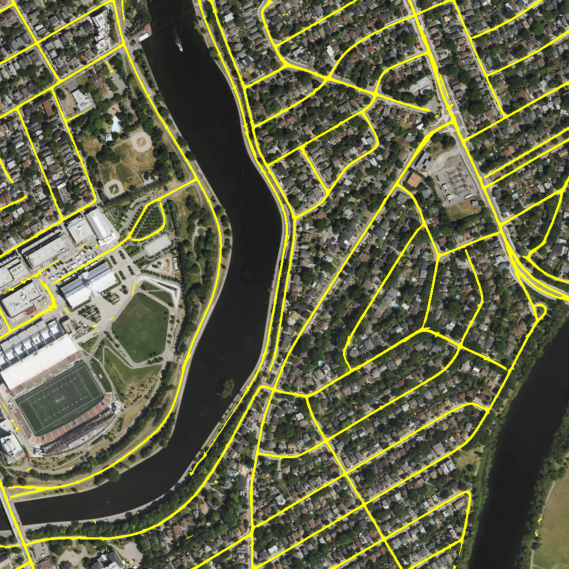}
\end{minipage}

\medskip

\includegraphics[width=0.23\linewidth]{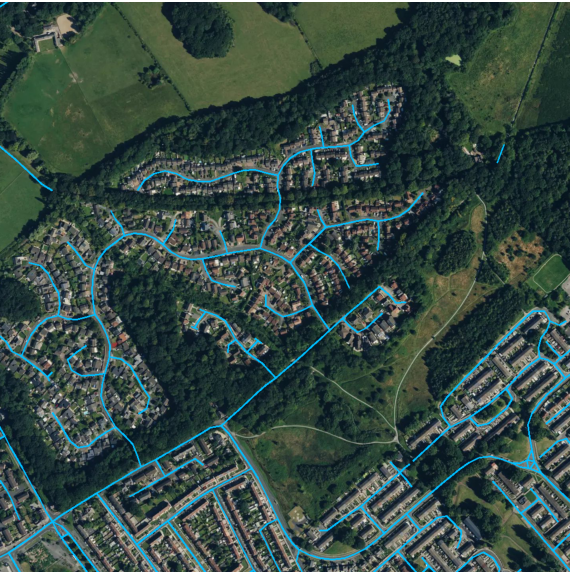}
\includegraphics[width=0.23\linewidth]{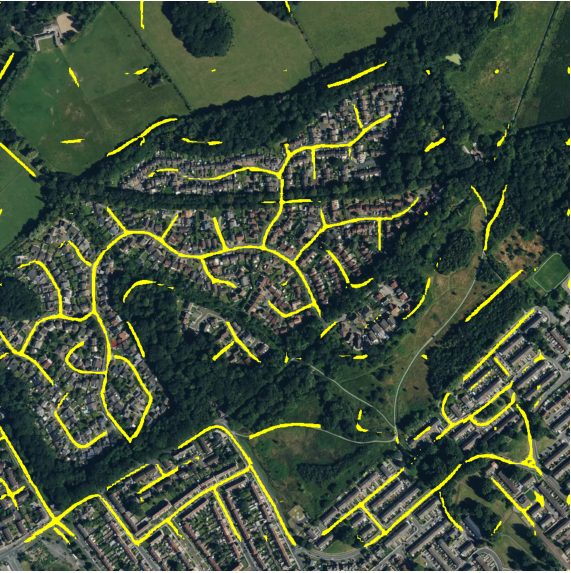}
\includegraphics[width=0.23\linewidth]{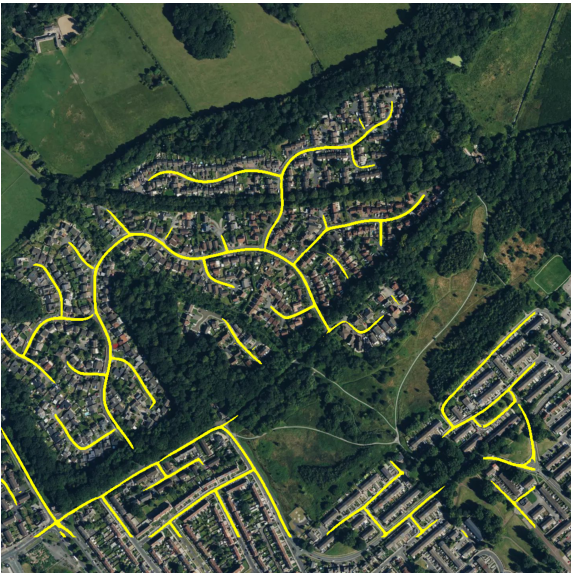}
\includegraphics[width=0.23\linewidth]{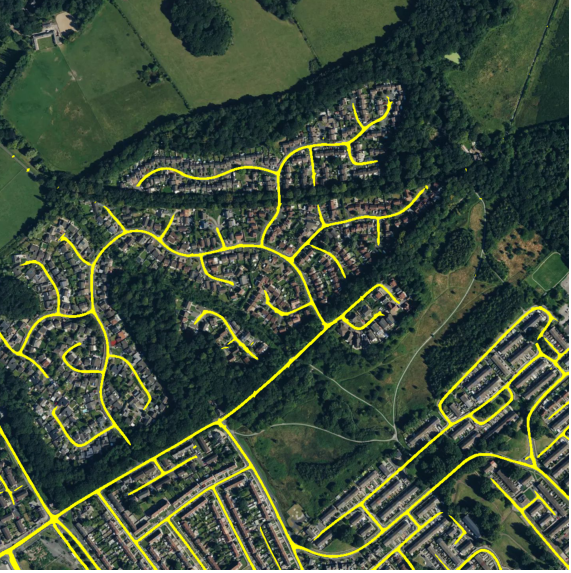}

\medskip

\includegraphics[width=0.23\linewidth]{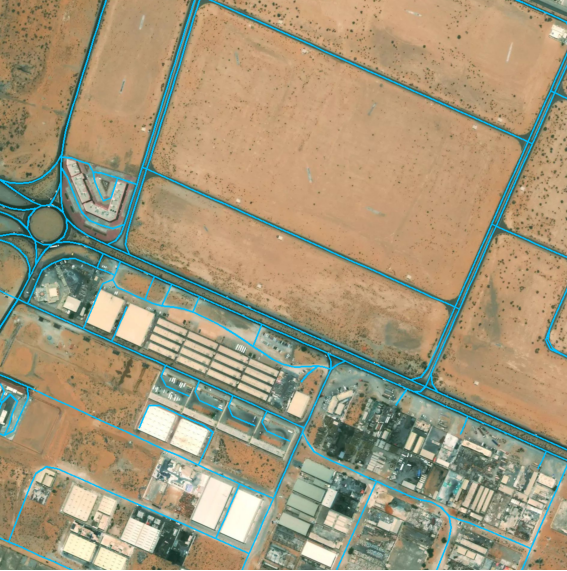}
\includegraphics[width=0.23\linewidth]{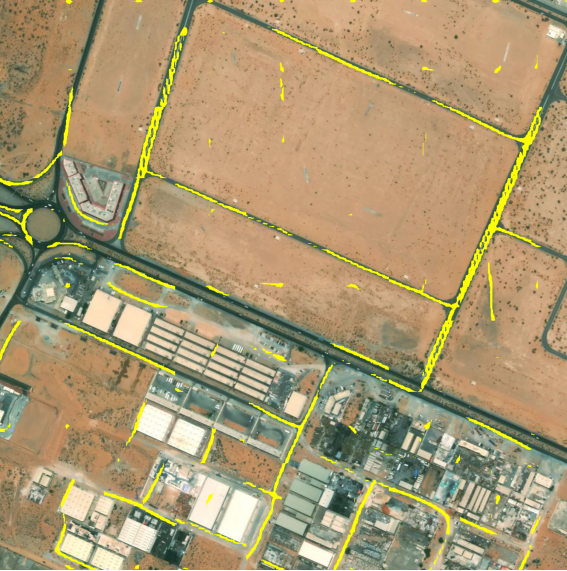}
\includegraphics[width=0.23\linewidth]{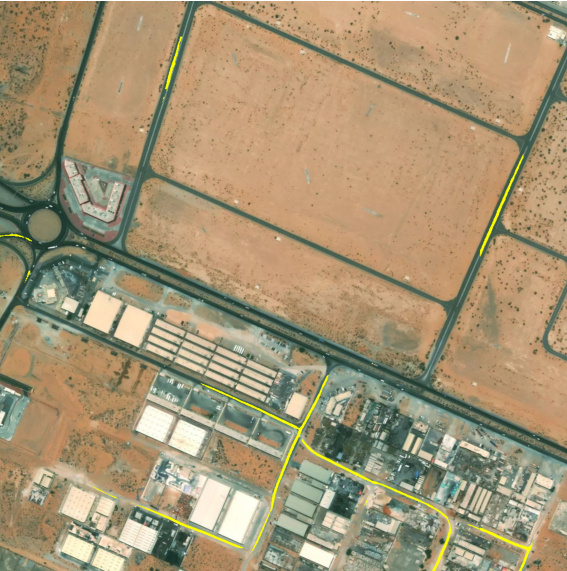}
\includegraphics[width=0.23\linewidth]{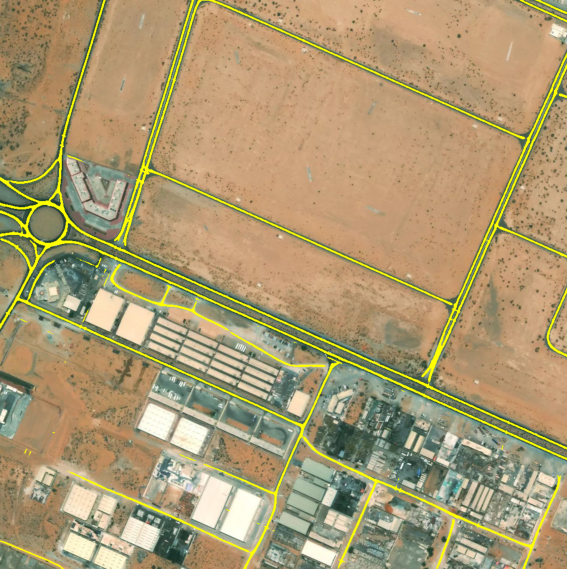}

\caption{
\textbf{Benchmark comparison on Global (Out-of-Domain) regions.}
Our model consistently preserves global connectivity and local road completeness across diverse unseen locations,
while baseline models (RNGDet++ and SAM) exhibit missing branches, fragmented paths, and broken intersections.
The improved continuity in our predictions demonstrates stronger generalisation to new geographies, road styles,
and imaging conditions, all without any domain-specific fine-tuning.
}
\label{fig:Seg_masks}
\end{figure}

A qualitative comparison of the segmented road networks from each method is shown in \cref{fig:Seg_masks}.
These results validate that the proposed continual-distillation framework transfers across sensors and geographies without per-benchmark fine-tuning.

The entire model contains only $31.1M$ trainable parameters and requires $134.3~GFLOPs$ per forward pass, making it substantially lighter than multi-stage baselines. On a single NVIDIA L40S GPU ($45~GB$ VRAM), full-resolution $1024\times1024$ inference averages $0.342~s$ per tile (${\sim}2.9$ images/s) with peak memory below $600~MB$. Training uses the AdamW optimiser with a learning rate of $2\!\times\!10^{-4}$ and a batch size of $32$ for $500$ epochs.

%% file: sec/7_ablation.tex
\section{Ablation Studies}
\label{sec:ablation}
\begin{figure}[H]
\centering
\begin{subfigure}[b]{0.32\linewidth}
    \includegraphics[width=\linewidth]{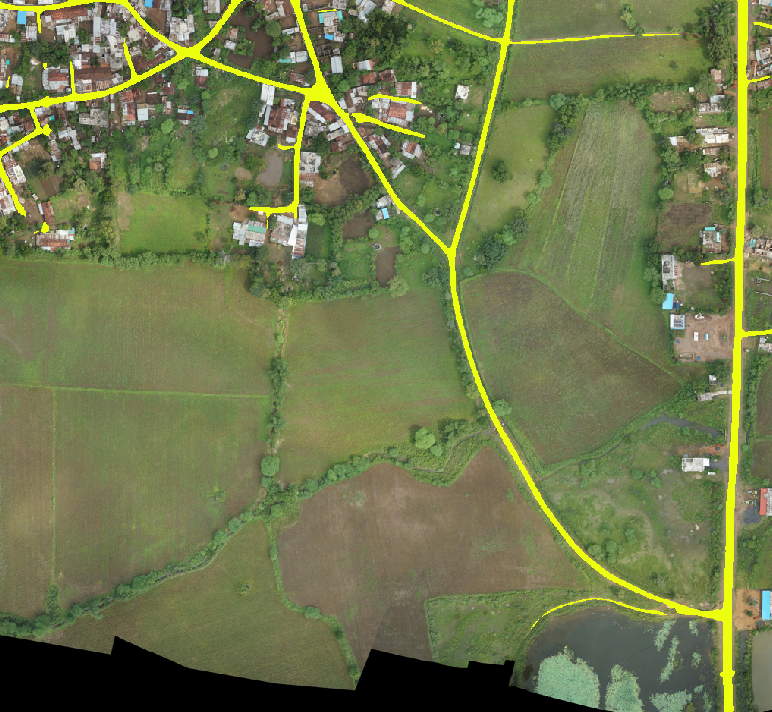}
    \label{fig:teacher}
\end{subfigure}
\hfill
\begin{subfigure}[b]{0.32\linewidth}
    \includegraphics[width=\linewidth]{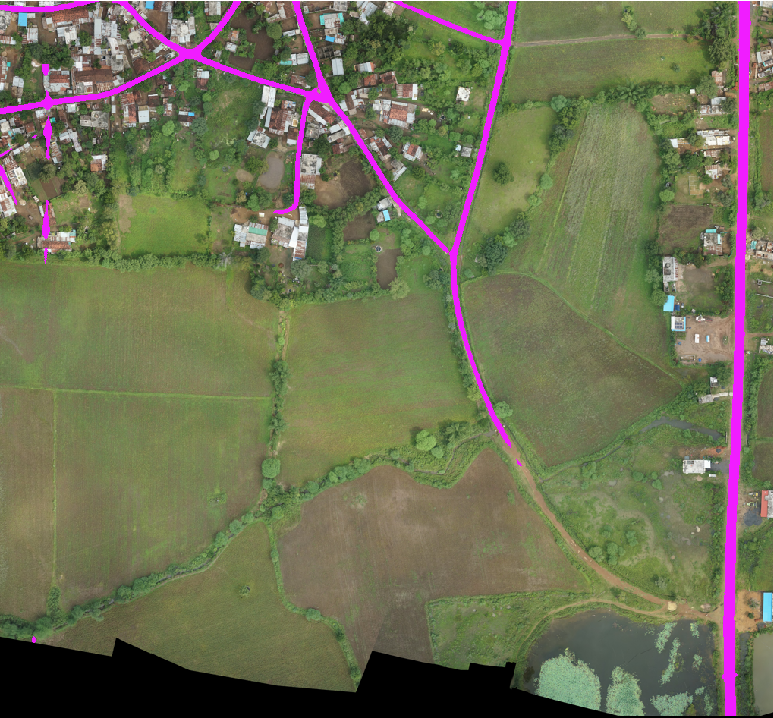}
    \label{fig:student_no_kd}
\end{subfigure}
\hfill
\begin{subfigure}[b]{0.32\linewidth}
    \includegraphics[width=\linewidth]{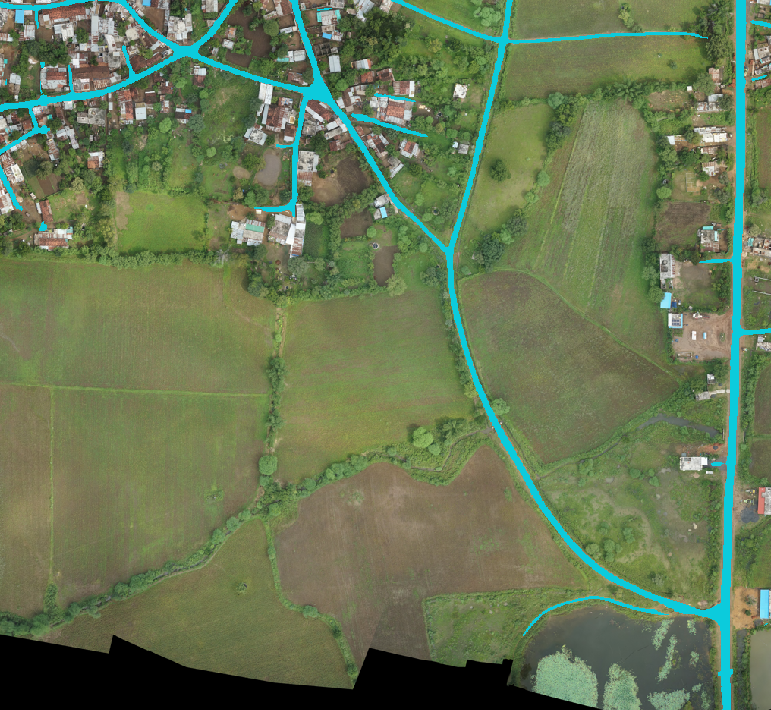}
    \label{fig:student_kd}
\end{subfigure}
\caption{Knowledge distillation mitigates catastrophic forgetting during low-resolution adaptation.
\textbf{(a)} The teacher, trained on high-resolution aerial imagery, recovers fine-grained road structure.
\textbf{(b)} A student initialised from the teacher's weights and adapted to lower-resolution data \emph{without} distillation degrades this structure, losing thin rural roads and subtle intersections as it overfits the new domain. \textbf{(c)} The distilled student, trained on the same lower-resolution data under supervision from the frozen teacher, preserves the teacher's structural priors while adapting.}
\label{fig:kd_contribution}
\end{figure}

\noindent\textbf{Contribution of knowledge distillation}
As shown in \cref{fig:kd_contribution}, removing Knowledge Distillation (KD) consistently reduces performance under the out-of-domain evaluation setting, leading to degradation in both F1 and APLS. Qualitatively, this appears as the loss of thin agricultural roads, subtle rural intersections, and fine structural details in unseen regions. In contrast, the KD-trained student preserves these structures while adapting to new domains.

\vspace{-0.25cm}
\vspace{-1cm}

\begin{figure}[H]
\centering
\begin{subfigure}[b]{0.48\linewidth}
    \centering
    \includegraphics[height=4cm]{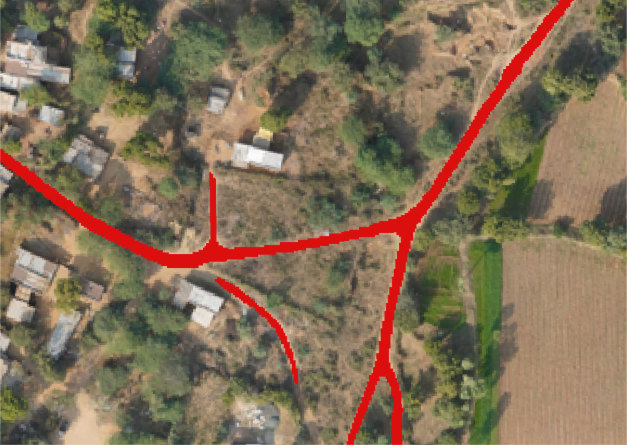}
    \caption{Without topology-preserving losses}
    \label{fig:without_skel}
\end{subfigure}
\hfill
\begin{subfigure}[b]{0.48\linewidth}
    \centering
    \includegraphics[height=4cm]{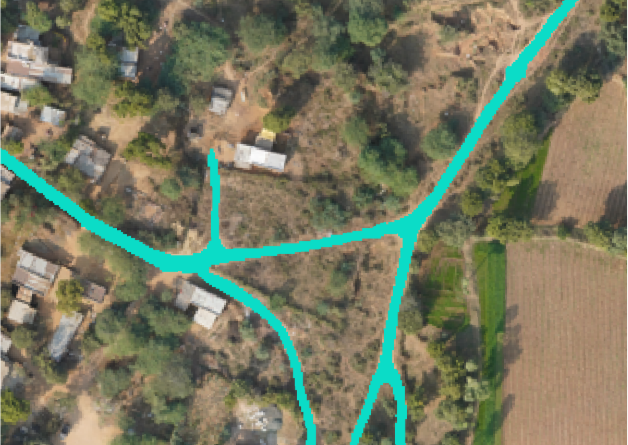}
    \caption{With topology-preserving losses}
    \label{fig:with_skel}
\end{subfigure}
\caption{Effect of the topology-preserving losses (skeleton-recall and intersection-aware).
\textbf{(a)} Trained without these losses, predictions exhibit micro-breaks along thin segments and fragmented, misaligned junctions.
\textbf{(b)} Adding both losses restores continuous centrelines and coherent intersection geometry, improving connectivity at T- and Y-junctions.}
\label{fig:ablation3}
\end{figure}

\vspace{-0.5cm}
\noindent\textbf{Contribution of topology-based losses}
Ablating individual loss components confirms their complementary roles as shown in \cref{fig:ablation3}. Removing the skeleton-recall loss reintroduces micro-breaks under vegetation and shadows. Omitting the intersection-aware loss yields fragmented Y-junctions and misaligned T-intersections. Excluding the focal loss causes faint rural tracks to vanish almost entirely. Each ablation produces a distinct and repeatable failure pattern, emphasising that all components are essential for maintaining network topology.

\begin{figure}[H]
\centering
\includegraphics[width=0.7\linewidth]{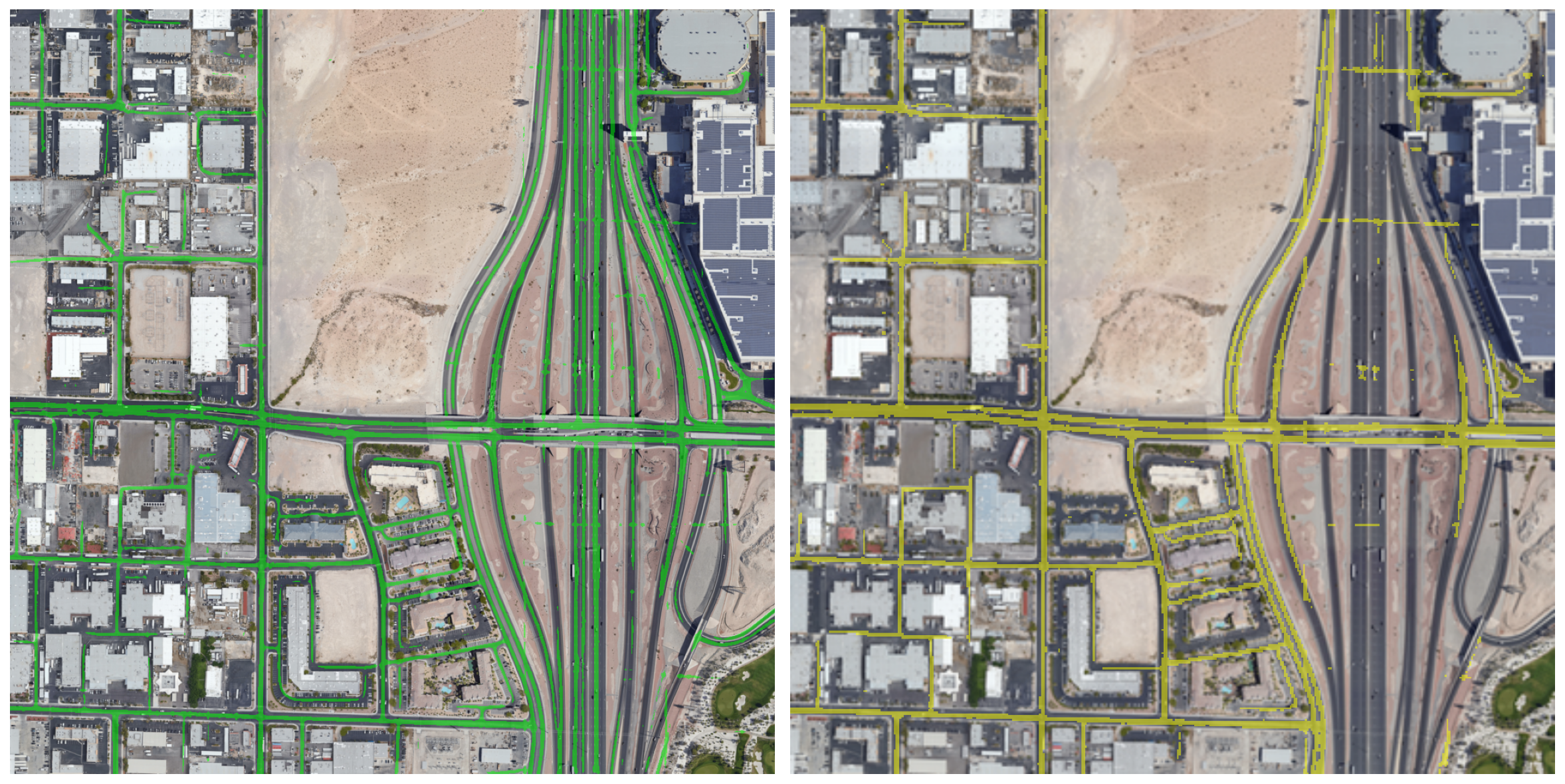}
\caption{\textbf{Effect of input patch size on segmentation quality.} With $1024\times1024$ patches (left), the model produces continuous, topologically coherent road networks, correctly resolving highway interchanges and side streets. Reducing the patch size to $400\times400$ (right) leads to fragmented predictions, missed road segments, and broken connectivity at complex intersections due to the reduced spatial context available to the encoder and dilation block.}
\label{fig:patch_size}
\end{figure}

\noindent\textbf{Effect of input patch size}
As shown in \cref{fig:patch_size}, reducing the input from $1024\times1024$ to $400\times400$ patches noticeably degrades segmentation quality: highway ramps become fragmented, side streets are missed, and intersection connectivity breaks down. Smaller patches limit the effective spatial context available to the dilation block, whose largest receptive field ($31\times31$ at rate $d{=}8$) spans a smaller ground area, leaving insufficient structural cues for resolving complex road layouts.

%% file: sec/8_failure.tex
\section{Future Work}
\label{sec:future}

\begin{figure}[H]
\centering
\begin{subfigure}[b]{0.48\linewidth}
    \includegraphics[width=\linewidth]{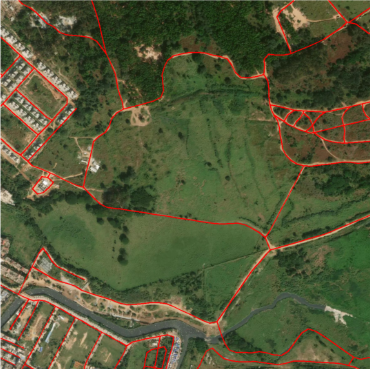}
    \caption{Ground Truth}
    \label{fig:failure_gt}
\end{subfigure}
\hfill
\begin{subfigure}[b]{0.48\linewidth}
    \includegraphics[width=\linewidth]{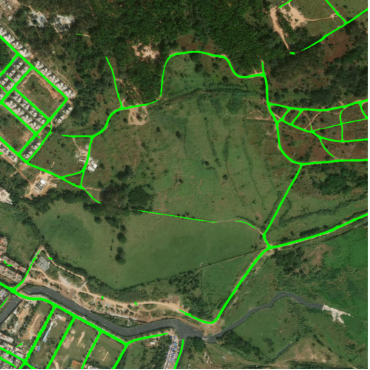}
    \caption{Prediction}
\end{subfigure}
\caption{\textbf{Representative Failure Cases of Our Proposed Model.} Thin unpaved roads and tree-occluded segments are still challenging due to low visibility and weak contextual cues, identifying key directions for future refinement.}
\label{fig:failure}
\end{figure}

The model generalises well across continents, sensors, and resolutions; however, several challenges remain. Extremely narrow rural tracks, often only one or two pixels wide and exhibiting minimal contrast with the surrounding terrain, may be partially missed due to a limited observable signal. In areas where dense tree canopies completely occlude the road surface, long segments become visually absent, requiring the model to infer structure solely from context, which can lead to minor thinning or short discontinuities even with skeleton-aware supervision. Under very low-resolution or degraded imagery, intersections lack sufficient geometric detail, resulting in softened or merged junctions that reflect input ambiguity rather than model failure. \Cref{fig:failure} illustrates representative conditions where the visual evidence for road topology is fundamentally constrained, highlighting the natural limitations of image-only inference.


Future work could incorporate multimodal cues, such as vector priors, to resolve cases where optical evidence is absent or ambiguous. Additionally, designing uncertainty-aware post-processing may further enhance robustness in occluded or low-resolution regions.

%% file: sec/9_conclusion.tex
\section{Conclusion}
\label{sec:conclusion}

This work introduced a simple and effective framework for road-network segmentation that improves robustness across resolutions, sensors, and geographic regions. By reframing the problem as one of representation stability rather than architectural novelty, we showed that combining curriculum-based cross-resolution distillation with topology-aware supervision enables a single model to generalise across diverse domains without per-domain fine-tuning. Across global in-domain, out-of-domain, and held-out City-Scale benchmarks, our method achieved consistent improvements over recent baselines, with the largest gains in cross-region and cross-dataset settings where prior models degrade most.
These results suggest that robustness in road extraction can be achieved through targeted training strategies, i.e, data curricula, distillation, and topology losses, rather than increasingly complex architectures. Our analysis is nonetheless bounded: the anchoring teacher is simultaneously high-resolution and single-region, so the benefit attributed to resolution cannot be fully disentangled from region-specific priors, and results are reported from single runs. Future work may address these factors with resolution-matched teachers and multi-run evaluation, and may extend the framework with deeper multi-stage curricula or multimodal cues such as vector priors to resolve cases where optical evidence is ambiguous.